\documentclass[runningheads]{llncs}
\usepackage{graphicx}
\usepackage{amsmath,amssymb} % define this before the line numbering.
\usepackage{color}
\usepackage{algorithm}
\usepackage{algpseudocode}
\usepackage{multirow}

\begin{document}
\pagestyle{headings}
\mainmatter

%===========================================================
\title{Simultaneous Adversarial Training - Learn from Others' Mistakes} % Replace with your title

\author{Zukang Liao}
\institute{$^{1}$Lite-On Singapore Pte. Ltd, $^{2}$Imperial College London}

\maketitle

%===========================================================
\begin{abstract}
Adversarial examples are maliciously tweaked images that can easily fool machine learning techniques, such as neural networks, but they are normally not visually distinguishable for human beings. One of the main approaches to solve this problem is to retrain the networks using those adversarial examples, namely adversarial training. However, standard adversarial training might not actually change the decision boundaries but cause the problem of gradient masking, resulting in a weaker ability to generate adversarial examples \cite{tramer2017space}. Therefore, it cannot alleviate the problem of black-box attacks, where adversarial examples generated from other networks can transfer to the targeted one. In order to reduce the problem of black-box attacks, we propose a novel method that allows two networks to learn from each others' adversarial examples and become resilient to black-box attacks. We also combine this method with a simple domain adaptation to further improve the performance.
\end{abstract}

%===========================================================
\section{Introduction}

Deep neural networks have been widely used to do facial recognition, surveillance, automotive driving and other tasks that require a high standard of safety. However, it has been found that by adding some unnoticeable perturbations to input images, i.e. adversarial noise, deep neural networks can be easily fooled \cite{goodfellow2014explaining}. Furthermore, these adversarial examples often transfer, which means that adversarial examples that fool one network can also easily fool another one. This is known as black-box attacks \cite{papernot2016practical}. In order to increase the robustness to the transferability of adversarial examples for faces, we propose a novel method that allows two networks to learn from each others' adversarial examples.

Standard adversarial training is proven to be effective to the same type of white-box attacks that are used to generate adversarial examples \cite{madry2017towards} \cite{szegedy2013intriguing}, but ineffective to black-box attacks due to gradient masking \cite{tramer2017ensemble}. After a network has been trained, its adversarial examples can be generated by various methods, such as Fast Gradient Sign Method or Least Likely Class method. These adversarial examples can easily fool the network as they are found using the network's weights and gradients, which is known as white-box attacks \cite{tramer2017ensemble}. Standard adversarial training retrains the network using original data and its adversarial examples to make it more robust to the same type of white-box attacks \cite{goodfellow2014explaining}. However, it has been found that adversarial examples generated from defended networks (using standard adversarial training) lose the ability to easily fool other undefended networks due to the problem of gradient masking \cite{tramer2017ensemble}. Furthermore, it has also been found that the decision boundaries of the defended networks (using standard adversarial training) remain unchanged, and thus the defended networks remain vulnerable to black-box attacks \cite{tramer2017space}.

We propose a method that trains two networks simultaneously to make both of them more resilient to black-box attacks from a third holdout network, and we call it Simultaneous Adversarial Training Method ($SATM$). $SATM$ is implemented and tested on the Adience dataset where 26,580 faces are labelled with gender and age group \cite{eidinger2014age}. We show some visually noticeable adversarial examples on Adience dataset and we find that, unlike databases for object recognition, those adversarial examples can be visually misleading to human beings when the adversarial noise is set to be large enough. Additionally, we find that, for networks that do not have batch normalisation layers \cite{ioffe2015batch} such as VGGNets \cite{simonyan2014very}, distribution of features of adversarial examples is different from that of original data. Therefore, we add a domain adaptation \cite{ganin2016domain} to further improve generalisability.

In this paper, we introduce methods to generate adversarial examples and methods to prevent white-box and black-box attacks in Section \ref{sec:related_work}. Dataset and some visually noticeable adversarial examples are shown in Section \ref{sec:database}. In Section \ref{sec:SATM} we introduce our methodology in details and in Section \ref{sec:exp} we show and analyse experiments results. Finally, we conclude what we have found and achieved and what can be done for future work in Section 6.

%------------------------------------------------------------------------- 

%===========================================================
\section{Related Work} \label{sec:related_work}

Various countermeasures for adversarial examples have been proposed for different types of attacks. Two main defense approaches are: 1) detecting and rejecting adversarial examples in testing stage to prevent adversarial attacks, and 2) making networks themselves more robust to adversarial examples, e.g. adversarial training. In this section, algorithms for generating adversarial examples are listed first, and then some state-of-the-art countermeasures are introduced.

%-------------------------------------------------------------------------
\subsection{Algorithms for Generating Adversarial Examples}
\subsubsection{Fast Gradient Sign Method (FGSM)}
FGSM computes one step gradient and adds the sign of the gradient to raw images \cite{goodfellow2014explaining}. It is defined as:
\begin{equation}
  x^{adv} = x +\varepsilon \cdot sign(\triangledown_{x}J(x, y_{true}))
  \label{FGSM_eq}
\end{equation}
where $x$ is raw images, $y_{true}$ is the labels, $J$ is the loss function and $\varepsilon$ controls the magnitude of the adversarial noise. Some variants of FGSM are used to enhance the generated adversarial examples. For example, Fast Gradient Value method removed the sign function \cite{rozsa2016adversarial} and \cite{dong2017boosting} combined momentum and an ensembling method with FGSM and won NIPS 2017 Targeted and Non-Targeted Adversarial Attacks Competition.

\subsubsection{Single-Step Least Likely Class method (Step-LL)}
Step-LL replaced $y_{true}$ with $y_{least-likely}$, and \cite{kurakin2016adversarial} showed it was the most effective for adversarial training on ImageNet. It is defined as:
\begin{equation}
  x^{adv} = x +\varepsilon \cdot sign(\bigtriangledown_{x}J(x, y_{least-likely}))
  \label{ILLC_eq}
\end{equation}
where $y_{least-likely} = argmin_{y}\left \{ p(y|x) \right \}$ and $p(y|x)$ is the probability that the network would predict $y$ given $x$.

%-------------------------------------------------------------------------
\subsubsection{Randomised single-step attack (R+Step-LL)}
\cite{tramer2017ensemble} showed that the vicinity of data points in loss function is not smooth. Simply using FGSM or ILLC might not suffice to find the actual adversarial direction. Therefore, they proposed a new randomised single-step attack which adds a small random step to escape from the non-smooth vicinity before computing gradients. R+Step-LL is defined as:
\begin{equation}
  x^{adv} = x{}' + (\varepsilon - \alpha )\cdot sign(\triangledown_{x{}'}J(x{}', y_{true})), \text{where } x{}' = x +\alpha \cdot sign(N(\mathbf{0}^{d}, \mathbf{I}^{d}))
  \label{RSSA}
\end{equation}

Other algorithms such as DeepFool \cite{moosavi2016deepfool}, CPPN EA Fool \cite{nguyen2015deep}, Hot/Cold method \cite{rozsa2016adversarial} and Natural GANs \cite{zhao2017generating} can also be used to generate adversarial examples. Particularly, stronger adversarial attacks that require much less perturbations such as $L_{2}, L_{0}, L_{\infty}$ attacks introduced by \cite{carlini2016towards} can be used as the main adversarial attacks for future work. They often succeed with $100\%$ probability with less than 4/256 distortion and normally they will not be visibly noticeable. Additionally, Step-LL, FGSM and R+Step-LL can be iterated many times but in this paper we would focus on single-step attack and we direct readers to \cite{gong2017adversarial} for further insights about iterative attacks. However, even though it is effective to weak iterative adversarial attacks, it has been broken by \cite{carlini2017adversarial}. We direct readers to a holistic survey \cite{yuan2017adversarial} for further information about generating adversarial examples.

In this paper, we use single-step R+Step-LL to generate adversarial examples for both training and testing with $\varepsilon = 16/256$ and $\alpha = \varepsilon/2$.

%-------------------------------------------------------------------------
\subsection{Countermeasures Without Adversarial Training}
Both the two main defense approaches mentioned above in Section \ref{sec:related_work} can fight against adversarial examples without adversarial training. In this section, we introduce cutting-edge methods of the first type and methods of the second type that do not involve adversarial training respectively. They were proven to be effective for weak adversarial attacks such as FGSM or Step-LL but some of them have been broken by stronger attacks \cite{carlini2017adversarial} such as $L_{2}, L_{0}, L_{\infty}$ attacks \cite{carlini2016towards}. 

In testing stage, adversarial examples can be prevented by either: 1) train a separate classifier to distinguish adversarial examples from clean data \cite{lu2017safetynet} \cite{metzen2017detecting} \cite{grosse2017statistical} \cite{hendrycks2017early}, or 2) find the differences between them by analysing their features. A wide variety of tricks can be used in the first case. For example, \cite{hosseini2017blocking} used soft labels and added a null class to counteract adversarial examples. In the second case, features that are chosen to distinguish adversarial examples from clean data vary. For example, \cite{feinman2017detecting} found the certainty of adversarial examples is higher than clean data from a Bayesian perspective and \cite{meng2017magnet} found coefficients in low-ranked components between adversarial examples and clean data were different. However, they have both been broken by stronger attacks with a slight increase in distortion \cite{carlini2017adversarial} \cite{carlini2017magnet}. Given that \cite{salimans2017pixelcnn++} found that adversarial examples have a different distribution from clean data, we combine a simple domain adaptation with our method to further improve the performance.

It is also possible to make networks more resilient to adversarial examples without adversarial training, \cite{papernot2016distillation} used high-temperature softmax to make models less sensitive to unnoticeable perturbations. \cite{simon2018adversarial} used double back-propagation to penalise large gradients and they found the regularisation scheme was equivalent to first order adversarial training. However, it has been shown that distillation does not make networks more robust to stronger attacks \cite{carlini2016towards} \cite{carlini2016defensive}. 

%-------------------------------------------------------------------------
\subsection{Adversarial Training Methods}
Using adversarial training introduced by \cite{goodfellow2014explaining} can prevent white-box attacks. However, instead of generally reducing adversarial vulnerability, the method can cause the problem of gradient masking \cite{tramer2017ensemble}. Due to the problem, after first-step adversarial training, the adversarially trained networks can only generate adversarial examples that are easier for undefended networks to classify, but the decision boundaries of the adversarially trained networks remain unchanged \cite{tramer2017space}. Thus, the adversarially trained networks remain vulnerable to black-box attacks.

In order to reduce the risk of black-box attacks, \cite{tramer2017ensemble} proposed a method of ensemble adversarial training which used one pre-trained network only for generating adversarial examples and then used those adversarial examples to train another network. This way, the adversarially trained network became more resilient to black-box attacks from a third holdout network.

Similarly, \cite{na2017cascade} proposed a method of cascade adversarial machine learning regularized with a unified embedding which uses one already defended network to generate adversarial examples to re-train another one. They found iterative attacks transfer more easily between networks that are trained using the same strategy i.e. standard training/Kurakin's adversarial training\cite{kurakin2016adversarial}. They also introduce a regularisation with a unified embedding which aligns features of adversarial examples and their corresponding clean data. This way, visually similar images would have similar features and it thus improves robustness of networks.

%===========================================================

\section{Adversarial Examples of Faces} \label{sec:database}

In this section, we first introduce the dataset we use and then we show comparisons between original data and visually noticeable adversarial examples. We find that, unlike object recognition, some adversarial examples can be misleading for human beings. Finally, we show some results on white-box attacks with different parameter values. Adversarial examples of faces are posing serious safety threads to many face recognition systems, driver monitoring systems and security surveillance systems. Therefore a more effective method to fight against this type of adversarial examples is necessary. 

The Adience dataset contains 26,580 unconstrained images of faces from 2,284 subjects, each of them is labelled with gender and eight age groups (0-2, 4-6, 8-13, 15-20, 25-32, 38-43, 48-53, 60-). These images were collected from the Flickr albums and they were authorisedly released by their
authors under the Creative Commons (CC) license. All images were taken completely ``in the wile", which means they were taken under different variations in appearance, noise, pose, blurring and lighting conditions \cite{eidinger2014age}. According to its protocol, five cross validation is used to make results more statistically significant \cite{levi2015age} \cite{liao2017local}.

\begin{figure}[h]
	\centering
	\includegraphics[height=2.5cm]{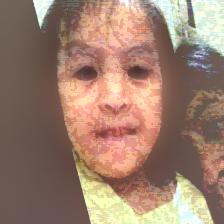}
	\includegraphics[height=2.5cm]{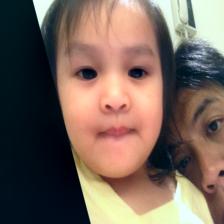}
	\includegraphics[height=2.5cm]{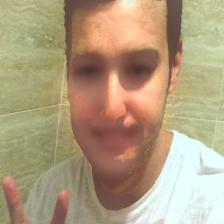}
	\includegraphics[height=2.5cm]{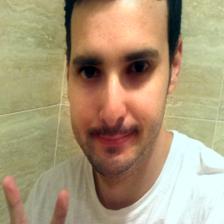}
	\includegraphics[height=2.5cm]{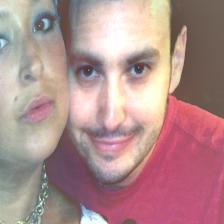}
	\includegraphics[height=2.5cm]{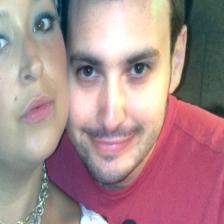}
	\includegraphics[height=2.5cm]{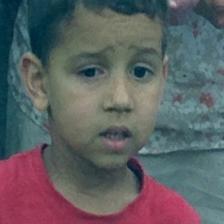}
	\includegraphics[height=2.5cm]{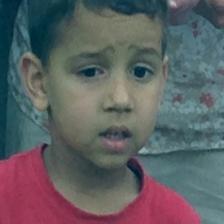}
	\caption{For each pair, the clean image is on the right and its adversarial example is on the left. Top left ($\varepsilon = 64/256$): labelled as the $1^{st}$ group but its adversarial example is misclassified to the $8^{th}$. Top right ($\varepsilon = 32/256$): labelled as the $5^{th}$ group but its adversarial example is misclassified to the $2^{nd}$. Bottom left ($\varepsilon = 16/256$): labelled as the $5^{th}$ group but its adversarial examples is misclassfied to the $7^{th}$. Bottom right ($\varepsilon = 8/256$): labelled as the $2^{nd}$ group but its adversarial example is misclassified to the $4^{th}$.}
	\label{fig:adversarial_examples}
\end{figure}

The comparison between some clean testing images and their adversarial examples on the Adience dataset is shown in Figure~\ref{fig:adversarial_examples}. These clean testing images are all classified correctly by a ResNet50-Face \cite{he2016deep} that is pre-trained on the VGGFace Database \cite{Parkhi15} and fine-tuned on the Adience dataset. However, their adversarial examples (that are generated using R+Step-LL) are all misclassified. Similar results can be found if we use VGG16-Face that is pre-trained on the VGGFace database. As shown in Figure~\ref{fig:adversarial_examples}, when $\varepsilon$ of the Equation~\ref{RSSA} (R+Step-LL) is set to be large enough, we can see the difference between the original data and adversarial examples clearly. We found that these adversarial examples can be visually misleading to human beings. For example, the adversarial example of the top left pair actually looks more senior than the clean one, and the adversarial example of the top right pair looks younger than the clean one.

\setlength{\tabcolsep}{5pt}
\begin{table}[h]
\centering
\caption{Classification rates of white-box attacks (R+Step-LL)}
\label{table:cr_adv_examples}
\begin{tabular}{cccccc}
\hline\noalign{\smallskip}
$\varepsilon$           & 32/256 & 16/256 & 8/256   & 4/256   & Clean data \\
\hline \noalign{\smallskip}
VGG16-Face  & 3.95\% & 4.30\% & 6.97\%  & 12.13\% & 56.73\%    \\ \hline \noalign{\smallskip}
ResNet-Face & 4.29\% & 8.79\% & 10.65\% & 21.8\%  & 59.08\%    \\ \hline
\end{tabular}
\end{table}
\setlength{\tabcolsep}{1.4pt}

When $\varepsilon$ is set to be between $8/256$ and $16/256$, adversarial examples become less visually noticeable but remain the ability to easily fool networks on the Adience dataset as shown in Table~\ref{table:cr_adv_examples}. We thus choose R+Step-LL with $\varepsilon = 16/256$ and $\alpha=\varepsilon/2$ to generate adversarial examples for the following experiments. 

%===========================================================
\section{Simultaneous Adversarial Training Method ($SATM$)}
\label{sec:SATM}
\begin{figure}[h]
	\centering
	\includegraphics[height=4cm]{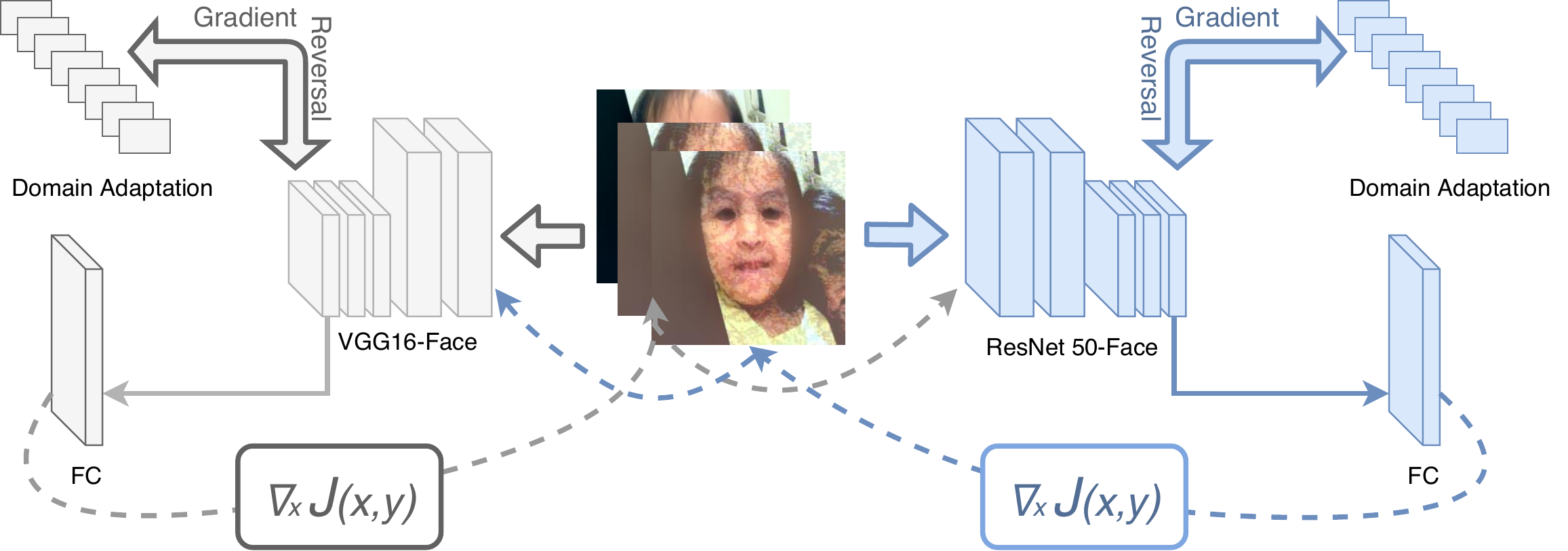}
	\caption{$SATM$ retrains two networks together using clean data and adversarial examples that are generated from the other network. Additionally, $SATM$ uses a domain adaptation to align features of clean data and adversarial examples, which improves generalisability.}
	\label{fig:SATM}
\end{figure}

SATM is proposed to alleviate black-box attacks. The procedure is shown in Figure~\ref{fig:SATM} and the algorithm is listed in Algorithm~\ref{alg:SATM}. This method re-trains two networks simultaneously using clean data and adversarial examples that are generated from the other network. This way, both networks become more resilient to black-box attacks from a third holdout network. We also combine domain adaptation with $SATM$, which improves generalisability especially for networks that do not include batch normalisation layers such as VGGNets.

\begin{algorithm}[h]
\caption{Simultaneous Adversarial Training Method ($SATM$)}\label{alg:SATM}
\begin{algorithmic}[1]
\State Fine-tune $Network^{A}$ and $Network^{B}$ using clean data
\Procedure{SATM}{$Network^{A}, Network^{B}$}
\State Initialise($\varepsilon$, $\alpha$, $lr$, $r_{adv}$, $r_{da}, Loss^{A}, Loss^{B}$)

\While{not early stopping} \Comment{Based on $Loss^{A}+Loss^{B}$}
\State $x_{adv}^{A} \gets $ R+Step-LL($Network^{A}$, $\varepsilon$, $\alpha$)
\State $x_{adv}^{B} \gets $ R+Step-LL($Network^{B}$, $\varepsilon$, $\alpha$)

\State $Loss^{A} \gets Network^{A}_{loss}(x_{clean}, y_{true})$
\State $\hspace*{18mm}+ r_{adv}*Network^{A}_{loss}(x^{B}_{adv}, y_{true})$
\State $\hspace*{18mm}+ r_{da}*DA^{A}(Network^{A}_{feature}(x_{clean}), Network^{A}_{feature}(x^{B}_{adv}))$

\State $Loss^{B} \gets Network^{B}_{loss}(x_{clean}, y_{true})$
\State $\hspace*{18mm}+ r_{adv}*Network^{B}_{loss}(x^{A}_{adv}, y_{true})$
\State $\hspace*{18mm}+ r_{da}*DA^{B}(Network^{B}_{feature}(x_{clean}), Network^{B}_{feature}(x^{A}_{adv}))$

\State $\omega^{A} \gets \omega^{A}+SGD(lr, Loss^{A}, \omega^{A})$
\State $\omega^{B} \gets \omega^{B}+SGD(lr, Loss^{B}, \omega^{B})$
\EndWhile
\EndProcedure
\bigskip
\Function{R+Step-LL}{$Network, \varepsilon, \alpha$}
\State $x{}' \gets x +\alpha \cdot sign(N(\mathbf{0}^{d}, \mathbf{I}^{d})$
\State $x_{adv} \gets x{}' + (\varepsilon - \alpha )\cdot sign(\triangledown_{x{}'}Network_{loss}(x{}', y_{true}))$
\State \textbf{return} $x_{adv}$ \Comment{Return adversarial examples}
\EndFunction
\end{algorithmic}
\end{algorithm}

\subsection{Simultaneous Adversarial Training}
$SATM$ re-trains two networks simultaneously but in this section we describe the method from the perspective of one network first, and then we introduce the scalability of $SATM$ and generalise it to training multiple (more than two) networks simultaneously.

As explained in Algorithm~\ref{alg:SATM}, for $Network^{A}$, $SATM$ uses clean data and adversarial examples that are generated from $Network^{B}$ to re-train it. In order to avoid the problem of gradient masking to the largest extent, we do not use adversarial examples that are generated from $Network^{A}$ itself (namely white-box adversarial examples) to re-train it. Therefore, after $Network^{A}$ has been re-trained using $SATM$, its adversarial examples still remain ``strong" enough to easily fool undefended networks. Similarly, after using $SATM$, $Network^{B}$'s adversarial examples also remain ``strong" enough. This means during the whole re-training process, $SATM$ uses clean data and ``strong" adversarial examples that are generated from $Network^{B}$ to re-train $Network^{A}$. If $Network^{B}$ is completely frozen, then $SATM$ becomes Ensemble Adversarial Training \cite{tramer2017ensemble} without using white-box adversarial examples. However, by using $SATM$, $Network^{B}$'s adversarial examples would not simply follow one distribution because $SATM$ also re-trains $Network^{B}$. This way, both networks can become more resilient to black-box attacks.

%-------------------------------------------------------------------------
\subsection{Domain Adaptation}
We combine domain adaptation with $SATM$ as shown in Figure~\ref{fig:SATM}. For networks without batch normalisation layers such as VGGNets, the distribution of clean data and the distribution of adversarial examples can be different. We use a simple domain adaptation block (a binary classifier with two fully-connected layers) to reduce the difference and improve generalisability. For $Network^{A}$, the domain adaptation block distinguishes features of clean data from features of $Network^{B}$'s adversarial examples. As shown in Figure~\ref{fig:SATM}, the gradient of the domain adaptation block will go though a gradient reversal layer and then flow back to $Network^{A}$. This way, those features generated from $Network^{A}$ become more indistinguishable and the generalisability is thus improved \cite{ganin2016domain}. More advanced domain adaptation methods such as Adversarial Discriminative Domain Adaption \cite{tzeng2017adversarial} can be used to replace the simplest domain adaptation block, but here we focus on $SATM$ and show that by combing a domain adaptation method with $SATM$, networks can be more resilient to black-box attacks.

%-------------------------------------------------------------------------
\subsection{$SATM$ with Multiple Networks}
Multiple (more than two) networks can be re-trained using $SATM$. If $SATM$ is used to re-train $\left \{ Network^{A}, Network^{B} ...Network^{K} \right \}$, for $Network^{A}$, clean data and adversarial examples that are generated from $Network^{B}, Network^{C} ...$ $Network^{K}$ would be used to re-train it. Adversarial examples can be generated by those networks interchangeably to ensure that we still use clean data 50 percents of the time. More advanced domain adaptation methods should be used to deal with the multi-domain adaptation problem. However, when more than two networks are included, the batch size can be smaller than two, which might affect the performance of batch normalisation layers and the domain adaptation method. Therefore we leave $SATM$ with multiple networks as a future work. We believe that $SATM$ with multiple networks would enumerate more types of adversarial perturbations. Therefore, networks might finally become more robust against black-box attacks using $SATM$ with multiple networks.

%===========================================================

\section{Experiments}
\label{sec:exp}
In this section we show experimental results on the Adience dataset using SATM. We use SATM to re-train fine-tuned VGG16-Face and ResNet50-Face simultaneously. We show results of white-box attacks first, and then we show results of black-box attacks both before and after using SATM and we show that SATM converges. Finally we show black-box attacks results from a third holdout network (which is chosen to be Resnet101 or
InceptionResNetV2 that are pre-trained on ImageNet and fine-tuned on the Adience dataset) before and after SATM. Results are evaluated using classification rate and one-off classification rate. One-off classification rate is defined as:
\begin{equation}
One\_off = \frac{\sum_{c=0}^{c=C-1} \sum_{k=max(c-1,0)}^{k=min(c+1, C-1)} confusion\_matrix[c, k]}{\sum_{c=0}^{c=C-1} \sum_{k=0}^{k=C-1} confusion\_matrix[c, k]}    
  \label{eq:oneoff}
\end{equation} where $C$ is the number of classes and $confusion\_matrix[c, k]$ is the number of examples of class $c$ and (mis-)classified as class $k$ \cite{levi2015age}. Five cross-validation that is defined in \cite{levi2015age} is used as the protocol to evaluate performance, so the results are statistically significant.

%-------------------------------------------------------------------------
\subsection{White-Box Attacks}
As shown in Table~\ref{table:white-box_results}, both networks become more robust to white-box attacks, even though $SATM$ is not designed to prevent white-box attacks; SATM does not use any white-box adversarial examples to re-train networks but the classification rate and one-off rate of white-box attacks still increase. As shown in Table~\ref{table:grey-box_results}, adversarial examples that are generated from these adversarially re-trained networks remain ``strong" enough to easily fool undefended networks such as InceptionV3 or InceptionResNetV2, which indicates that the improvement for white-box attacks does not come from the problem of gradient masking.

\setlength{\tabcolsep}{5pt}
\begin{table}[h]
\centering
\caption{Results of white-box attacks. Classification rate and One-off increase after using $SATM$}
\label{table:white-box_results}
\begin{tabular}{ccc}
\hline \noalign{\smallskip}
                          & Classification rate      & One-off  \\ 
\hline
VGG16-Face                & 4.30\% $\rightarrow$ \textbf{7.22\%}  & 13.90\% $\rightarrow$ \textbf{21.43\%}  \\ \hline
ResNet50-Face             & 8.79\% $\rightarrow$ \textbf{12.56\%}  & 26.30\% $\rightarrow$ \textbf{29.89\%}  \\ \hline
\end{tabular}
\smallskip
\smallskip
\smallskip

\centering
\caption{The first column is the models we are testing on, and the second column is the models that are used to generate adversarial examples (attacks). After using $SATM$, classification rate and one-off almost remain unchanged.}
\label{table:grey-box_results}
\begin{tabular}{cccc}
\hline  \noalign{\smallskip}
                          & Adv model            & Classification rate      & One-off  \\ \hline
InceptionV3    & VGG16-Face                & 14.49\%  $\rightarrow$ \textbf{14.32\%} & 45.01\% $\rightarrow$ \textbf{47.60\%}\\ \hline
InceptionV3    & ResNet50-Face             & 23.03\% $\rightarrow$ \textbf{19.67\%} & 58.81\% $\rightarrow$ \textbf{53.99\%}  \\ \hline
\end{tabular}
\end{table}
\setlength{\tabcolsep}{1.4pt}

%-------------------------------------------------------------------------
\subsection{Black-Box Attacks}
As shown in Table~\ref{table:black-box_results}, after VGG16-Face and ResNet50-Face have been adversarially trained using $SATM$, they both become resilient to each others' adversarial examples. Classification rates of black-box attacks increase to $52.40\%$ and $43.25\%$, which are very close to the classification rates on clean testing data ($56.29\%$ and $51.77\%$).

\setlength{\tabcolsep}{5pt}
\begin{table}[h]
\centering
\caption{Results of black-box attacks before and after using $SATM$.}
\label{table:black-box_results}
\begin{tabular}{cccc}
\hline \noalign{\smallskip}
                          & Adv model                 & Classification rate      & One\_off \\
\hline \noalign{\smallskip}
VGG16-Face                & ResNet50-Face             & 35.39\% $\rightarrow$\textbf{52.40\%} & 66.99\% $\rightarrow$ \textbf{88.55\%}  \\ \hline
ResNet50-Face             & VGG16-Face                & 16.25\% $\rightarrow$ \textbf{43.25\%} & 36.16\% $\rightarrow$ \textbf{76.11\%}  \\ \hline
\end{tabular}
\end{table}
\setlength{\tabcolsep}{1.4pt}
%Adv trained VGG16-Face    & ResNet50-Face             & 35.39\% & 66.99\%  \\ \hline
%Adv trained ResNet50-Face & VGG16-Face                & 25.58\% & 49.96\%  \\ \hline

%-------------------------------------------------------------------------
\subsection{Convergence}

\begin{figure}[h]
	\centering
	\includegraphics[height=3.2cm]{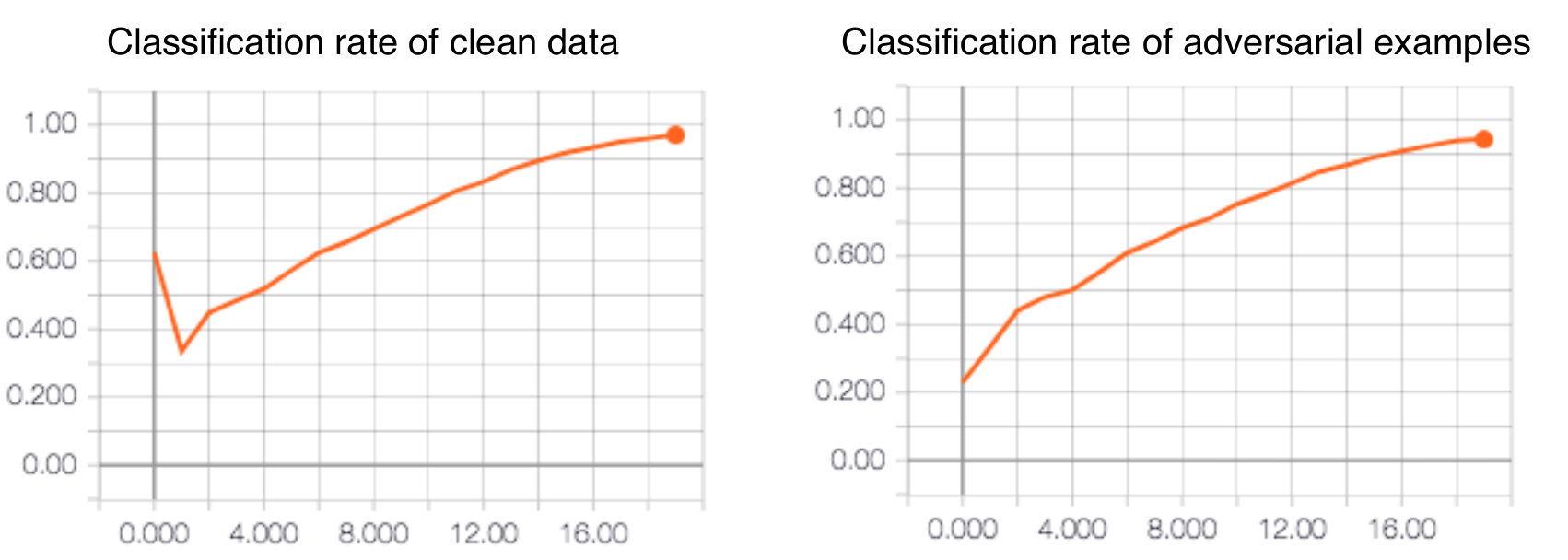}
	\caption{VGG16-Face's classification rates of clean data and adversarial examples on training set. X asix is the number of epochs.}
	\label{fig:tr_adv}
	\includegraphics[height=3.2cm]{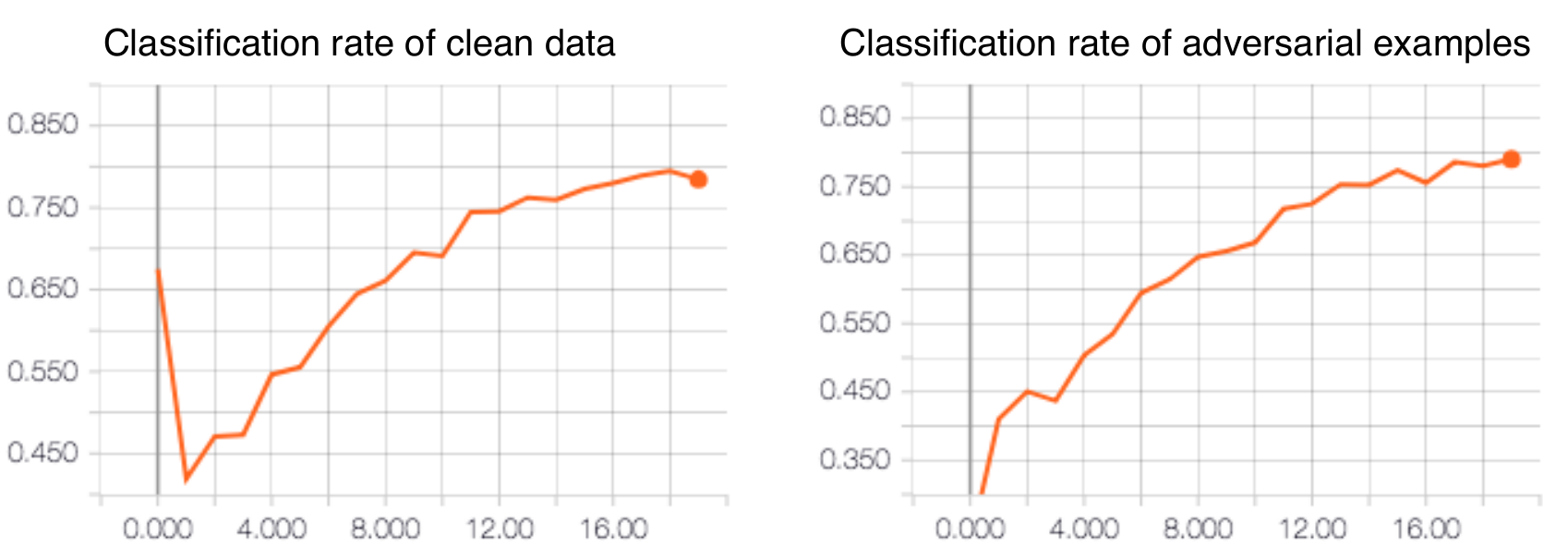}
	\caption{VGG16-Face's classification rates of clean data and adversarial examples on validation set. X asix is the number of epochs.}
	\label{fig:val_adv}
\end{figure}

As shown in Figure~\ref{fig:tr_adv} and Figure~\ref{fig:val_adv}, during training, the classification rates of clean data and adversarial examples are relatively similar to each other both on training set and validation set. After using $SATM$, the classification rates of clean data and adversarial examples of training set are both around $90\%$, which indicates $SATM$ converges. In another word, after a certain number of iterations (in this case after 10,000 iterations with mini-batches of size 8 and learning rate of $e^{-5}$), the distributions of adversarial examples change more slowly than the networks adapt to the change. We run experiments on a Tesla K80 GPU and $SATM$ converges in 20 hours.

\subsection{Black-Box Attacks from a Third Holdout Network}
As shown in Table~\ref{table:black_holdout}, we can see a significant improvement of the performance on black-box attacks from a third holdout networs (ResNet101 and InceptionResNetV2). We also show that $SATM$ outperforms the state-of-the-art method (Ensemble Adversarial Training) on the Adience database. In order to directly compare $SATM$ with Ensemble Adversarial Training, we do not use any white-box examples to re-train the networks, and we find that the best performances all come from $SATM$. For InceptionNets, When $\varepsilon$ is set to $16/256$, both undefended and adversarially trained VGG16 and ResNet50 are resilient to their adversarial examples. Therefore we set $\varepsilon$ to $32/256$ when testing on InceptionResNetV2.

\setlength{\tabcolsep}{4.5pt}
\begin{table}[h]
\centering
\caption{Classification rates/one-off rates of black-box attacks from a third holdout network, i.e. InceptionResNetV2 or ResNet101-Img which are pre-trained on ImageNet and fine-tuned on the Adience dataset.}
\label{table:black_holdout}
\begin{tabular}{ccccc}
\hline \noalign{\smallskip}
              & Adv model     & Undefended  & Ensemble          & SATM     \\
\hline \noalign{\smallskip}
VGG16-Face    & ResNet101-Img & 31.29/62.61\% & 35.87/68.04\%   &\textbf{37.40/72.16\%} \\ \hline
ResNet50-Face & ResNet101-Img & 22.59/47.57\% & 25.84/54.62\%   &\textbf{29.19/58.27\%} \\ \hline
VGG16-Face    & InceptionResNetV2 & 37.74/62.48\% & 39.61/66.31\% &\textbf{42.59/67.23\%} \\ \hline
ResNet50-Face & InceptionResNetV2 & 40.23/71.09\% & 43.12/78.62\% &\textbf{48.26/80.63\%} \\ \hline

\end{tabular}
\end{table}

%-------------------------------------------------------------------------
\subsection{Experiments on Networks with the Same Structure}
We set both $Network^{A}$ and $Network^{B}$ to be ResNet50-Face (the same structure and the same initialisation), and found that this led to divergence of the algorithm. A potential reason is: this way, these two networks are learning from their own ``white-box attacks", while they are not able to mask each others' gradients. Additionally, we also conduct experiments on ResNet50-Face and ResNet50-ImageNet (the same structure and different initialisation). As shown in Table~\ref{same_structure}, this combination leads to an $2.42\%$ accuracy improvement on adversarial examples generated by ResNet101-Img, however it also leads to a $2.64\%$ accuracy decrease on adversarial examples generated by InceptionResNetV2. This may be because that adversarial examples generated by ResNet101-Img resembles adversarial examples generated by ResNet50-Img, while they are less correlated with adversarial examples generated by InceptionResNetV2.

\setlength{\tabcolsep}{4.5pt}
\begin{table}[h]
\centering
\caption{Comparison (Classification rates/one-off rates) between training ResNet50-Face with VGG16-Face and training ResNet50-Face with ResNet50-Img.}
\label{same_structure}
\begin{tabular}{cccc}
\hline \noalign{\smallskip}
              & Adv model     & VGG16-Face  & ResNet50-Img     \\
\hline \noalign{\smallskip}
ResNet50-Face & ResNet101-Img  & 29.19/58.27\% &  \textbf{31.61/61.94\%} \\ \hline
ResNet50-Face & InceptionResNetV2 & \textbf{48.26/80.63\%} & 45.62/77.54\% \\ \hline

\end{tabular}
\end{table}

\subsection{Experiments on Other Databases}
A series of experiments on MNIST and ImageNet database are also conducted, however, no significant improvement on ImageNet database and no improvement on MNIST database (not worse either) are found compared with ensemble adversarial training method \cite{tramer2017ensemble}. For ImageNet, we use the same testing method as \cite{tramer2017ensemble} where 10,000 testing images are randomly chosen. InceptionResV2 and VGG16-Face are trained using SATM, and ResNet101 is chosen to be the third holdout network to generate black-box attacks. As shown in Table~\ref{compare_imagenet}, compared with ensemble adversarial training, SATM decreases top-1 error rate by $0.6\%$ and top-5 error rate by $0.2\%$ on ImageNet. For MNIST, we re-trained structure $A$ and $B$ \cite{tramer2017ensemble}, and we report averaged black-box attacks error rate for structure $A$. As shown in Table~\ref{compare_imagenet}, no significant improvement can be found on MNIST database using SATM compared with ensemble adversarial training.

\setlength{\tabcolsep}{4.5pt}
\begin{table}[h]
\centering
\caption{Classification error rate on ImageNet and MNIST using ensemble adversarial training and SATM.}
\label{compare_imagenet}
\begin{tabular}{|l|c|c|c|l|}
\hline
         & \multicolumn{2}{c|}{ImageNet}       & \multicolumn{2}{c|}{\multirow{2}{*}{MNIST}} \\ \cline{2-3}
         & Top 1         & Top 5 & \multicolumn{2}{c|}{}                       \\ \hline
Ensemble & 27.0\%           & 7.9\%            & \multicolumn{2}{c|}{5.2\%}                  \\ \hline
SATM     & 26.4\%           & 7.7\%            & \multicolumn{2}{c|}{5.6\%}                  \\ \hline
\end{tabular}
\end{table}

Different adversarial training methods can be more effective in specific domains (hand-written numbers, faces or objects classification). A potential reason can be that the distribution of adversarial examples of faces changes more quickly during the re-training process. Therefore, by using SATM, networks would have the chance to learn from more adversarial examples with different distribution. However, this needs to be supported by a more complete hypothesis and further experiments and we leave this topic for future work.

%===========================================================
\section{Conclusion and Future Work}
We propose a novel method ($SATM$) which trains multiple networks simultaneously to improve their robustness to black-box attacks without encountering the problem of gradient masking. In order to achieve this, $SATM$ uses adversarial examples that are generated from other networks to re-train the targeted network. This way, these networks learn from others' adversarial examples dynamically and thus all become more resilient to single-step black-box attacks. Furthermore, we also include a simple domain adaptation method to align features of clean data and features of adversarial examples to improve the performance. We conduct a series of experiments and show that, by using $SATM$, networks become slightly more resilient to single-step white-box attacks and significantly more resilient to single-step black-box attacks, while their adversarial examples remain ``strong" enough to easily fool undefended networks. We also show that $SATM$ outperforms the state-of-the-art method on single-step black-box attacks from holdout networks. In order to further improve the performance, white-box examples can be used with $SATM$ in various ways, stronger iterative adversarial attacks can be used, and a more deliberate domain adaptation method can be combined with $SATM$ for future work.

\newpage

%===========================================================
\bibliographystyle{splncs}
\bibliography{egbib}

%this would normally be the end of your paper, but you may also have an appendix
%within the given limit of number of pages
\end{document}